\documentclass[letterpaper]{article} % DO NOT CHANGE THIS
\usepackage[]{aaai23}  % DO NOT CHANGE THIS
\usepackage{times}  % DO NOT CHANGE THIS
\usepackage{helvet}  % DO NOT CHANGE THIS
\usepackage{courier}  % DO NOT CHANGE THIS
\usepackage[hyphens]{url}  % DO NOT CHANGE THIS
\usepackage{graphicx} % DO NOT CHANGE THIS
\usepackage{natbib}  % DO NOT CHANGE THIS AND DO NOT ADD ANY OPTIONS TO IT
\usepackage{caption} % DO NOT CHANGE THIS AND DO NOT ADD ANY OPTIONS TO IT
\usepackage{algorithm}
\usepackage{algorithmic}

\usepackage{fitch}

\usepackage{enumerate}
\usepackage{amsthm}

\usepackage{newfloat}
\usepackage{listings}
\DeclareCaptionStyle{ruled}{labelfont=normalfont,labelsep=colon,strut=off} % DO NOT CHANGE THIS
\floatstyle{ruled}
\newfloat{listing}{tb}{lst}{}
\floatname{listing}{Listing}
\title{Extending Logical Neural Networks Using First-Order Theories}
\author {
    Aidan Evans,
    Jorge Blanco
}
\affiliations {
    aidan.evans@yale.edu, jorge.blanco@yale.edu\\
    Yale University
}

\usepackage{bibentry}
\begin{document}

\maketitle

\begin{abstract}
Logical Neural Networks (LNNs) are a type of architecture which combine a neural network's abilities to learn and systems of formal logic’s abilities to perform symbolic reasoning. LLNs provide programmers the ability to implicitly modify the underlying structure of the neural network via logical formulae. In this paper, we take advantage of this abstraction to extend LNNs to support equality and function symbols via first-order theories. This extension improves the power of LNNs by significantly increasing the types of problems they can tackle. As a proof of concept, we add support for the first-order theory of equality to IBM's LNN library and demonstrate how the introduction of this allows the LNN library to now reason about expressions without needing to make the unique-names assumption.
\end{abstract}

\section{Introduction}\label{sec:intro}

    A current new and promising neural network architecture, Logical Neural Networks (LNNs) proposed by \cite{lnns}, has shown promising results as an architecture which combines neural network's abilities to learn and systems of formal logic's abilities to perform symbolic reasoning. LNNs work by associating with each neuron in the network a subexpression of a real-valued logic formula, such as weighted \L{}ukasiewicz logic. Because of the one-to-one association of neurons to logical formulae, LNNs have the ability to represent their decisions in terms of logic; therefore, unlike other neural architectures, we have the ability to easily interpret the decisions of LNNs while still retaining the robust learning ability of neural networks. In summary, LNNs allow for a neural architecture with explainable decisions.
    
    Moreover, because of LNNs' inherit ``two-sided'' nature -- i.e., their one-to-one correspondence of formulae to neurons -- LNNs provide programmers the ability to modify the underlying structure of the neural network without needing to actually work with the neurons themselves. In other words, LNNs allow programmers to work with neural networks at an abstracted level via easy to understand and concise logical formulae. Thus, LNNs can be seen to have two \textit{sides}: a ``logic side'' and ``neuron side''. This notion of abstraction is akin to that inherit in the design of the Internet because the Internet was made such that one could use and design Internet applications without needing to worry about the low-level details and protocols actually used to transmit data. In this paper, we argue that this abstraction allows for the robust design of more complex extensions of LNNs without needing to work with ``low-level'' neurons. 
    
    LNNs operate by taking as input a \textit{knowledge base} where each entry is a logical formula. Currently, LNNs support formulae expressed in first-order logic (FOL) -- a very powerful language with respect to what one can express in it. FOL languages \textit{typically include} predicate, variable, constant, and function symbols -- along with an equality operator represented as a special predicate or logical constant symbol. The current architectural design of LNNs, however, only supports the use of expressions in FOL \textit{without equality and functions symbols}. While equality and functions are not needed to obtain the full expressive ability of FOL, they nevertheless provide one with the ability to write and reason with formulae in a much \textit{easier and natural way}. Therefore, in this work, in order to demonstrate the robustness of LNNs, we extend LNNs to support equality and function symbols.
    
    Furthermore, in doing so, we restrict ourselves to working at the logic side of LNNs -- demonstrating the power of abstraction inherit in LNNs. Specifically, we introduce both equality and functions as \textit{first-order theories}, i.e., additional axioms expressed in terms of and reasoned with FOL. Therefore, we need only introduce these axioms into the network via logical formulae during the construction of the network. In this work, we explain how the introduction of first-order theories increases the domain of problems LNNs can reason about. We additionally provide a description of what the introduction of these theories corresponds to in terms of the low-level neuron side of LNNs. Finally, as a proof of concept, we introduce support for equality into the IBM's LNN Python library\footnote{\url{https://github.com/IBM/LNN}}, therefore, allowing IBM's system to now reason about the equality symbol. 
    
    \iffalse
    The introduction of first-order theories to LNNs opens the doors for the wide-ranging application of LNNs in automated theorem proving (ATP) tasks. ATP systems determine whether a given conjecture follows from a specified set of axioms expressed in a logical system. Because LNNs currently do not support equality and function symbols, the domain of ATP problems solvable by LNNs is extremely small as most FOL theorems use these symbols. Since we give LNNs the new ability to reason about equality and function symbols, we then evaluate the reasoning ability of LNNs by investigating LNNs' abilities on a wider range ATP tasks than previously possible.
    \fi
    
    % Roadmap  
    In this paper, we first provide a background on LNNs in Section \ref{sec:back}; we focus primarily on the basic structure of an LNN, i.e., how symbols and neurons interact and the connections between real-valued logic and activation functions. We also briefly discuss how inference and learning work in LNNs. In Section \ref{sec:ext}, we introduce the theoretical framework used to add equality and function to LNNs and discuss how the addition of equality and functions using this framework affects the neural structure of LNNs. In Section \ref{sec:impeval} we discuss the details of our implementation of equality into IBM's LNN library and provide an example of how this increases the domain of problems LNNs can handle. In Section \ref{sec:conc} we conclude and discuss future work.

\section{Background}\label{sec:back}

In this section, we provide a brief introduction to the LNN framework created by \cite{lnns}.

\subsection{Logical Neural Networks}\label{sec:lnn}

Historically, the field of artificial intelligence has focused on either \textit{statistical AI} or \textit{symbolic AI}. Statistical AI for example includes the study of neural networks, while symbolic AI has included the study of ``good old-fasioned'' AI, i.e., deductive systems. Statistical AI allows for inductive reasoning which enables the model to generalize inferences from a set of given data; symbolic AI allows for deductive reasoning which enables the model to draw for-sure conclusions using formal systems of logic. Moreover, symbolic AI allows for one to have a clear, explainable sequence of reasoning steps -- statistical AI tends to just be a black-box with no way to understand \textit{why} the model made the decision it did. LNNs aim to create a bridge between these two distinct approaches and, therefore, allows a model to perform both inductive and deductive reasoning -- taking advantage of the benefits of both the statistical and symbolic AI approaches.

\begin{figure}[t]
    \centering
    \includegraphics[scale=0.25]{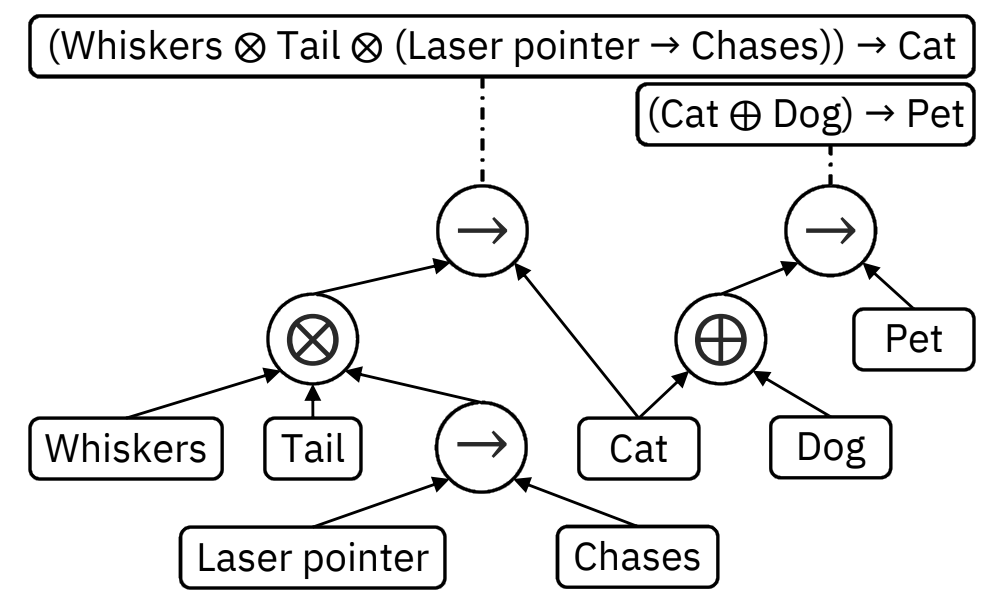}
    \caption{Example Network \cite{lnns}}
    \label{fig:ex1}
\end{figure}

\paragraph{Basic Structure of LNNs}
Currently, the framework of LNNs proposed by Riegel et al. combine the capabilities of neural networks and FOL by associating neurons with subformulae of real-valued logic, specifically, \L{}ukasiewicz logic. Overall, the structure of the neural network associated with the formula is exactly the formula's syntax tree; an example is shown in Figure \ref{fig:ex1}.

 In \L{}ukasiewicz logic, instead of having logical expressions evaluate to simply \textbf{true} or \textbf{false}, they now evaluate to some real-valued number between 0 and 1; we call this number the \textit{truth value}. We define a \textit{threshold of truth}, $\frac{1}{2} < \alpha \leq 1$, such that we consider a statement \textbf{true} if its truth value is above $\alpha$ and \textbf{false} if below $1 - \alpha$.   

\begin{figure}[t]
    \centering
    \includegraphics[scale=0.2]{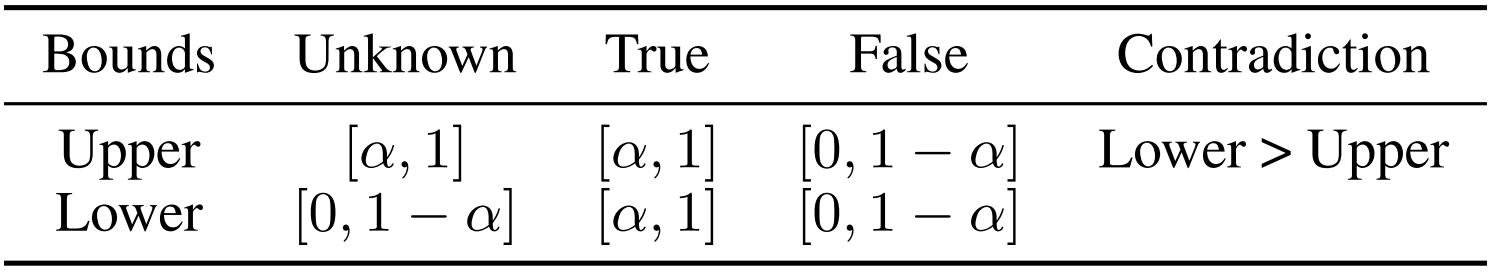}
    \caption{Primary Truth Value Bound States \cite{lnns}}
    \label{fig:bounds}
\end{figure}

Each neuron returns a pair of numbers between 0 and 1 called the \textit{lower} and \textit{upper} bounds. These numbers determine which \textit{primary state} the neuron is in; these states are displayed in Figure \ref{fig:bounds}.

Essentially, each binary logical connective of the input formula, such as conjunctions ($\wedge$ or $\otimes$) or disjunctions ($\vee$ or $\oplus$), is associated with a neuron; the neuron's activation function is the real-valued logical operation, such as the following for \L{}ukasiewicz logic, which computes the \textit{truth-value} of the operation:
\begin{align*}
    \text{Conjunction:}&\ \  p \otimes q=\max \{0, p+q-1\}\\
    \text{Disjunction:}&\ \  p \oplus q=1-((1-p) \otimes(1-q))=\min \{1, p+q\}\\
    \text{Implication:}&\ \  p \rightarrow q=(1-p) \otimes q=\min \{1,1-p+q\}
\end{align*}

These activation functions are then generalized to a ``weighted real-valued logic'' which allows one to express the importance of a subformula; these weights are updated via backpropagation during learning. The unary negation ($\neg$) connective and existential ($\exists$) and universal ($\forall$) quantifiers are represented as \textit{pass-through nodes} which are neurons with basic unweighted activation functions. For example, negation is simply: $\neg p = 1 - p$.

FOL predicates are represented as input neurons where each predicate has an associated table of \textit{groundings}, i.e., our known input data for what the predicate should evaluate to given various inputs. The input to the predicates of the table are the FOL constants, which are symbols associated with objects within our \textit{domain of discourse}, i.e., what we're talking about.

\paragraph{Inference}
To perform inference, LNNs perform an \textit{upward} and \textit{downward} pass through the network. The upward pass propagates the truth values from the predicates through the network, calculating the truth value of the entire formula itself. The downward pass works using the believed truth value of the entire formula to compute the truth values of its subformulae and predicates. Specifically, the upward pass will compute truth bounds for each subformula and, ultimately, the entire formula using the truth bounds of the predicates; the downward pass will then tighten these bounds for each subformula and, ultimately, each predicate until convergence. Convergence is guaranteed for propositional logic (i.e., no quantifiers and only nullary predicates) but is not guaranteed for FOL due to FOL's undecidability.

\begin{figure*}[t]
    \centering
    \includegraphics[width=\textwidth]{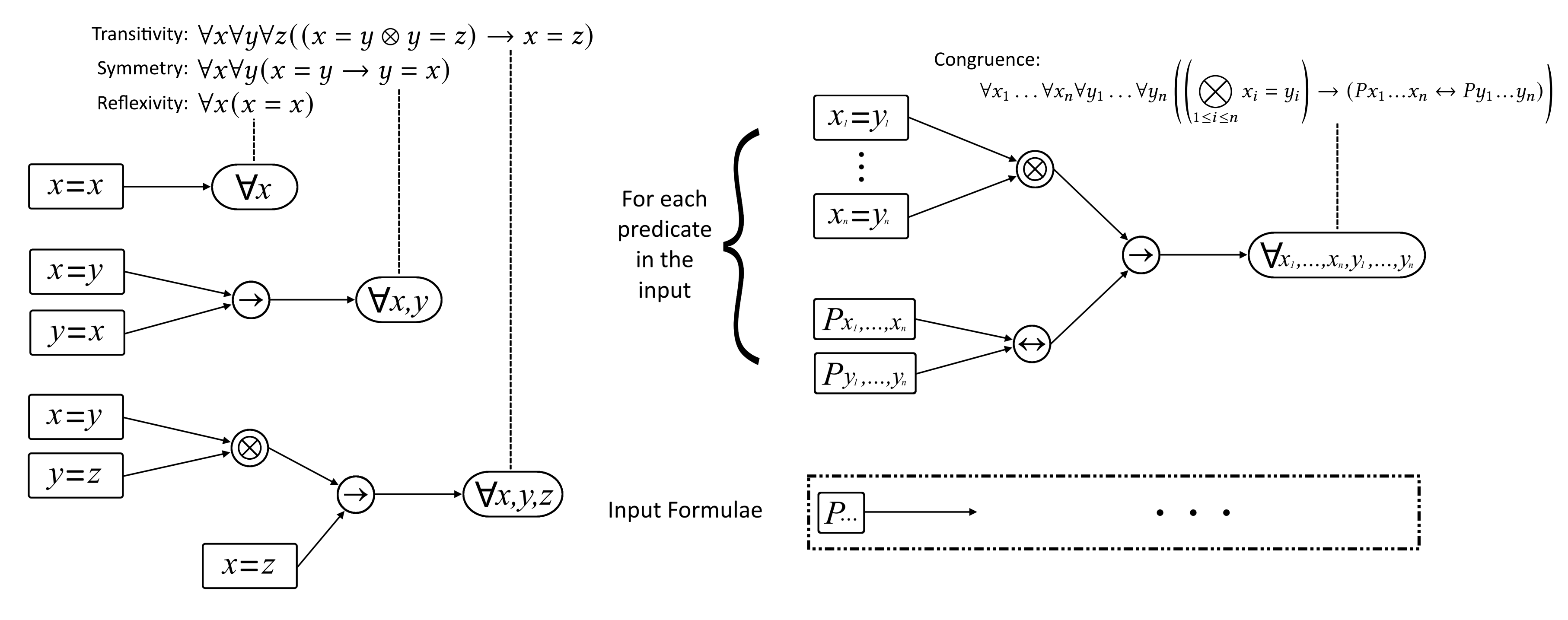}
    \caption{Equality Model Schematic}
    \label{fig:eqschem}
\end{figure*}

\paragraph{Learning}
Because we are working with real-valued formulae, the equations used are differentiable and, therefore, we are able to use backpropagation to update the parameters, such as the weights of each connective or the truth value bounds, of the formulae. The loss function for the optimization via backpropagation aims to minimize the amount of contradiction present in the model; in other words, it aims to remove as many contradictory conclusions as possible. This allows LNNs to learn from noisy and conflicting data sets.

\section{Theoretical Results: Extending LNNs via First-Order Theories}\label{sec:ext}
 
In this section, we describe how to formalize equality and functions as first-order theories and how adding these theories into an LNN will affect its underlying structure.

%we discuss how we may extend what LNNs can reason about by the incorporation of first-order theories into an LNN. 
Informally, a first-order theory is a set of symbols which we may include in our FOL formulae such that these symbols also have some sort of \textit{additional meaning}. A set of axioms also included with the theory specify the additional meaning placed on these symbols. Essentially, first-order theories allow us to formalize more complex structures and concepts in which we would like to discuss: e.g., equality, functions, lists, arrays, trees, etc. In our case, we wish to formalize equality and functions.

\subsection{Equality}\label{sec:eqtheory}

\subsubsection{Equality in FOL} We first wish to include a theory of equality to formalize the notion of the equality operation. The theory of equality is a common first-order theory and is a popular way to introduce an equality symbol to FOL.\footnote{cf. \cite{calccomp}} To do so, we begin by introducing the equality symbol itself as a binary predicate: ``=''. Specifically, we will add a sort of ``universal predicate'' to the LNN which is defined from the beginning of the LNN model and, therefore, always accessible for use in any formula; we discuss the detailed implementation of this in practice in Section \ref{sec:impeval}.
Meanwhile, this predicate is supposed to capture our notion of what it means for two things to be equal. Specifically, in FOL, the equality operator should be comparing two \textit{terms}. A term in FOL is either a constant, variable, or the application of a function to another term. Because we currently do not have functions, we restrict our discussion in this subsection to terms which are simply constants or variables. 

In FOL, formulae are evaluated under a specific \textit{interpretation} or \textit{model}. Terms are then said to \textit{refer} to some \textit{object} in the domain of the interpretation; under different interpretations, different terms may refer to different objects. Within an interpretation, we use the equality symbol to say that two terms refer to the same object. For example, if we had two constants, $a$ and $b$, and we then said that $a = b$, this means that both $a$ and $b$ refer to the same object within our domain.

Using this knowledge, we then add axioms which formalize the intended meaning we wish the equality symbol to have. Because we wish the equality symbol to say that two terms are the same if the refer to the same object, we are essentially claiming that the equality symbol is an equivalence relation: it's partitioning our terms into classes which all refer to the same object. Thus, the predicate we use to represent equality should be reflexive, symmetric, and transitive. We, thus, have our first three axioms:

\begin{align*}
    \forall x &(x = x) \tag{Reflexivity}\\
    \forall x \forall y &( x = y \rightarrow y = x) \tag{Symmetry}\\
    \forall x \forall y \forall z &((x = y \otimes y = z) \rightarrow x = z) \tag{Transitivity}
\end{align*}

We also desire that two equal terms can be replaced by one another; i.e., we can substitute any term for one to which it's equal. For example, say we had some unary predicate, $P$, and terms $x$ and $y$ such that $x = y$. Then, we'd wish to say that if $Px$, then $Py$ too since $x = y$. Here, because $x = y$, we should be able to replace any occurrences of $x$ with $y$.

In the case of our unary predicate, $P$, we may state this with the following axiom:
\begin{align*}
    \forall x \forall y (x = y \rightarrow (Px \leftrightarrow Py))
\end{align*}
This now ensures that we can replace equal terms with each other in our expressions for $P$. 

More generally, say that $P$ is an $n$-ary predicate. We would then include the following axiom:

\newcommand\mydots{\makebox[1em][c]{.\hfil.\hfil.}}
\begin{align*}
    &\forall x_1 \dots \forall x_n \forall y_1 \dots \forall y_n \\
    &\left(\left(\bigotimes_{1 \leq i \leq n} x_i = y_i \right) \rightarrow (Px_1 \mydots x_n \leftrightarrow Py_1 \mydots y_n) \right) \tag{Congruence}
\end{align*}

\subsubsection{Effect on Neural Structure}
In order to add these axioms to our LNN model, we may add our first three axioms (reflexivity, symmetry, and transitivity) during the instantiation of our model since they only make a claim about our equality predicate, which is also introduced from the beginning; i.e., we add the three axioms as formulae of our knowledge base from the very start. This will lead our neural network to contain neurons corresponding to each axiom's formula.

In the case of our fourth axiom (congruence), however, we must handle this for every predicate; therefore, whenever a new predicate is introduced into our knowledge base, we also add the appropriate congruence axiom for this predicate, which also leads to the introduction of neurons for each of the congruence axioms.

We include a model schematic displaying the effect on the network from adding these axioms in Figure \ref{fig:eqschem}.

%For now, we turn to discussing how adding these axioms to an LNN affects the underlying structure of the LNN's neural network.

\subsection{Functions}\label{sec:functheory}

For the incorporation of functions as a first-order theory, we had much less to go off of. The introduction of equality as something added ``later on'' to FOL is a much more common practice than for functions. For example, in the proof of G\"odel's Completeness Theorem, we first prove the claim for FOL without equality and then prove that introducing equality does not change anything. Normally, however, functions are assumed to be included in FOL from the start -- unlike equality. Therefore, for the formalization of functions, we did not have many prior resources to work with. We describe our results for formalizing functions as a first-order theory below.

\paragraph{Functions as ``Functional Relations''}
To formalize functions as a first-order theory, we must understand what functions are fundamentally. Now, an $n$-ary function is used to map $n$ inputs to an output such that no set of inputs can map to two different outputs. Moreover, both functions and predicates are fundamentally relations; the only difference being that the relation for a function is constrained to capture how each input may only map to one output. Formally, we say that an $(n+1)$-ary relation, $R$, is \textit{functional} if and only if
\begin{align*}
    \forall w_1 \dots \forall w_n \forall x \forall y ((Rw_1\mydots w_nx \otimes Rw_1\mydots w_ny) \rightarrow x = y) \tag{Functional}
\end{align*}
This condition ensures that if the first $n$ inputs to our $(n+1)$-ary relation are the same, then so will the $(n+1)^{th}$ input. In other words, we may say that this ensures that the first $n$ inputs map to the value of the $(n+1)^{th}$ input as we desire for functions. Thus, we may say that for any $n$-ary function, $f$, we can construct an $(n+1)$-ary functional relation, $R_f$, such that for every term $x$ and terms $w_1$ through $w_n$,
\begin{align*}
    R_fw_1\mydots w_n x \text{\ \ if and only if\ \ } f(w_1, \mydots, w_n) = x \tag{$\star$}
\end{align*}
Since predicates are simply relations, we can then rewrite each function as a predicate.
    
\paragraph{Rewriting Formulae to use Functional Relations instead of Functions}
Furthermore, when it comes to how functions may be used in FOL formulae, functions are another type of term -- like constants and variables. Notice that for any predicate $P$ and term $t$, $Pt$ is logically equivalent to $\exists x(t=x \otimes Px)$; see Appendix \ref{app:funcproof} for a proof. In other words, we may extract the term out of our predicate and introduce an existentially quantified variable which must be equal to our term. 

Therefore, by $(\star)$ and in the case of $t$ being a function $f(t_1\mydots{}t_n)$, we may say that $Pt$ is logically equivalent to $\exists x (R_ft_1\mydots{}t_n x \otimes Px)$. Since the relation $R_f$ may be represented simply as a predicate, we have now rewritten an FOL formula which contained functions in terms of an FOL formula without any functions. Since functions may themselves have functions as arguments, this procedure repeats until all the functions are removed. 

\paragraph{Recursively Extracting Functions} To explain the recursive extraction of functions, we do so by providing a basic example of the phenomena. Say we have a unary predicate $P$, two unary functions $f$ and $g$, and a constant $c$. Let $R_f$ and $R_g$ be the functional relations associated with $f$ and $g$, respectively. 

Now say we wish to remove the use of all functions from the formula $Pf(g(c))$. Since $f(g(c))$ is a term of $P$, we first extract this out so that $P$ does not contain any terms which are functions:\[
    Pf(g(c)) \Rightarrow \exists x (R_fg(c)x \otimes Px)
\]
Note that after rewriting, however, we are still left with a predicate containing a term which is a function. While $P$ may be function free, $R_f$ is not. Therefore, we again apply the rewriting rules to our new formula in order to again extract any functions from the terms of our newly introduced predicate $R_f$:\[
    \exists x (R_fg(c)x \otimes Px) \Rightarrow \exists x (\exists y (R_gcy \otimes R_fyx) \otimes Px)
\]
Thus, we now have an equivalent function-free formula for $Pf(g(c))$.\\

Therefore, in order to add functions to LNNs, we may simply rewrite the formulae of an LNN's knowledge base to remove functions via the rewriting procedure outlined above and then add axioms ensuring that the predicates we introduced for each function during the rewriting steps are functional.

\subsubsection{Effect on Neural Structure}

\begin{figure*}
    \centering
    \includegraphics[width=\textwidth]{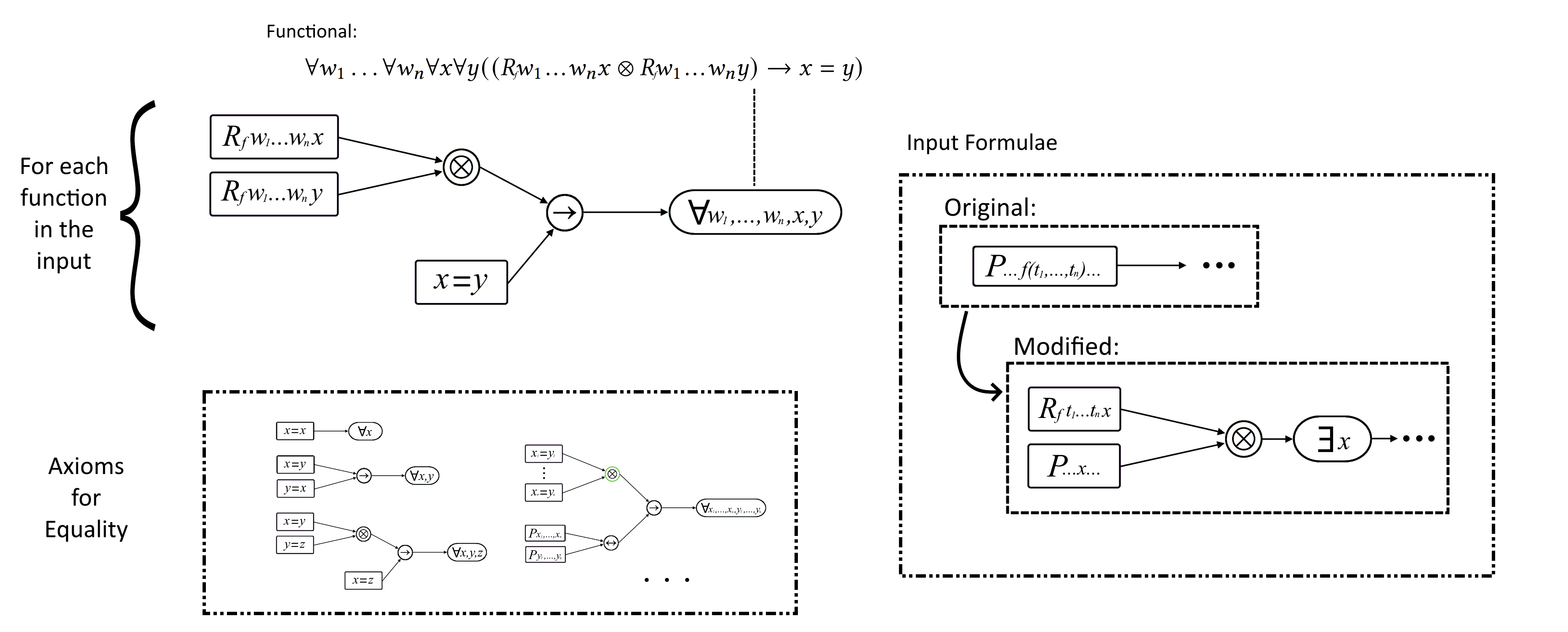}
    \caption{Function Model Schematic}
    \label{fig:funcschem}
\end{figure*}

Similar to equality, we will add axioms for each of the predicates associated with a function; specifically, we will do so for those predicates we wish to be defined as functional. This will involve introducing several new neurons to the network to handle the axiom for each function. We present a model schematic displaying the effect on the network from adding these axioms in Figure \ref{fig:funcschem}.

We will also modify the neuronal structure corresponding to our input formula as well. Specifically, for each predicate which contains a function as an argument, we will end up adding additional neurons for each of the functions and neurons for existential quantification and conjunction since those operators are introduced from the rewriting process.

\section{Proof of Concept: Implementation of Equality}\label{sec:impeval}

In this section, we include details of our implementations of extending IBM's LNN library to include equality and provide an example demonstrating how the introduction of equality expands the domain of problems LNNs can now represent.
Our implementation of equality is available in an online repository.\footnote{\url{https://github.com/nsnave/LNN}}

\subsection{Description of Implementation}\label{sec:eqimp}

First, to use equality within a model, we specify that an LNN model instance should have support for equality during its instantiation:
\begin{lstlisting}[numbers=none]
from lnn import Model
model = Model(theories=["equality"])
\end{lstlisting}
This ensures that equality's required axioms are added from the start.

In IBM's LNN library, predicates are instances of a \texttt{Predicate} class. Since we are treating equality as a predicate, to add an equality operator to IBM's LNN library, we define a new variable\footnote{``variable'' here is being used to describe a variable in Python -- not to be confused with our earlier discussion of variables within FOL.} \texttt{Equals} which is an instance of the \texttt{Predicate} class; this variable is then added as an additional import like the other operators. Therefore, users have access to the variable globally.

This also allows for the \texttt{Model} class to have access to the equality variable so we can add the axioms we need to for equality within the \texttt{Model} class. During the instantiation step above, this introduces the axioms for reflexivity, symmetry, and transitivity. We add the congruence axioms for each predicate when each predicate is added with the \texttt{add\_predicates} function. We now discuss the implementation via an example.

\begin{listing}[tb]%
\caption{Example listing {\tt test\_same\_name.py}}%
\label{lst:listing}%
\begin{lstlisting}[language=Python]
from lnn import (Model, Equals)

model = Model(theories=["equality"])
dog = model.add_predicates(1, "dog")

model.add_facts({
    dog.name: {
        "Aggie": Fact.TRUE
    },
    "equals": {
        ("Aggie", "Fruton"): Fact.TRUE
    }
})

model["query"] = Not(dog("Fruton"), world=World.AXIOM)

model.infer()
print(model["query"].state())
\end{lstlisting}
\end{listing}

\subsection{Example}
This full example is presented in Listing \ref{lst:listing} and in the online repository.\footnote{\url{https://github.com/nsnave/LNN/tests/reasoning/logic/fol/test_same_name.py}} We now walk through it step by step.

Say we wish to introduce a unary predicate \texttt{dog} which tells us whether the input argument refers to a dog. We would then write the following:
\begin{lstlisting}
from lnn import (Model, Equals)

model = Model(theories=['equality'])
dog = model.add_predicates(1, 'dog')
\end{lstlisting}
At this point, the model now has axioms for reflexivity, symmetry, and transitivity (via line 3) along with congruence for the \texttt{dog} predicate (via line 4).  

If we then wish to say that there is a dog named ``Aggie'' and that Aggie's nickname is ``Fruton'', we'd introduce the following facts to the model:
\begin{lstlisting}[numbers=none]
model.add_facts({
    dog.name: {
        "Aggie": Fact.TRUE
    },
    "equals": {
        ("Aggie", "Fruton"): Fact.TRUE
    }
})
\end{lstlisting}
This establishes that \texttt{dog(`Aggie')} and \texttt{`Aggie' = `Fruton'} are true statements. We can then demonstrate our model's ability to reason about equality by having it prove that \texttt{dog(`Fruton')} must also be true. To do so, we introduce as an axiom that \texttt{Not(dog(`Fruton'))} is true; if this results in our model reporting that we have a contradiction, then we know that \texttt{Not(dog(`Fruton'))} must in reality be false and, therefore, \texttt{dog(`Fruton')} is true.

To accomplish this, we begin by first declaring that \texttt{Not(dog(`Fruton'))} is an axiom of our model:

\begin{lstlisting}[numbers=none]
model["query"] = Not(dog('Fruton'), world=World.AXIOM)
\end{lstlisting}

Then, we'll have the model deduce what it can about its knowledge base via the \texttt{infer} function and print the resulting state of our query, which ultimately reports a contradiction:
\begin{lstlisting}[numbers=none]
model.infer()
print(model['query'].state())
\end{lstlisting}

Currently, LLNs necessarily make the \textit{unique-names assumption}; this assumes that every constant refers to a unique object in the domain. Therefore, in the example above, there would be no way to declare that both the names ``Aggie'' and ``Fruton'' refer to the same dog. Now that we've introduced equality, we need not make this assumption as the above example proves. Because of our introduction of equality, we were able to state that both the constants ``Aggie'' and ``Fruton'' do in fact refer to the same object. Thus, we were able to deduce that whatever ``Fruton'' referenced was a dog because we knew that what ``Aggie'' referenced was a dog. Therefore, the introduction of equality as a first-order theory greatly expanded what LNNs are able to reason about without actually changing how the underlying architecture's neurons work; we only had to add or modify the logical formulae of the network.

\subsection{A Note on Functions}\label{sec:funcimp}

In general, the implementation of functions would be significantly more complicated since it involves the actual rewriting of the input functions.
In order to add support for functions to IBM's library, we would need to rewrite the formulae added to the model; however, IBM's current implementation of LNNs is not friendly to rewriting formulae. A better approach for introducing theories which involve rewriting the input formulae would be to modularize the rewriting process by introducing a separate parser. This parser would take as input a simple abstract syntax tree for each input formulae, apply the rewriting rules, and then pass these rewritten formulae to IBM's LNN library as usual.

\section{Conclusion}\label{sec:conc}

% 
%
% - conclusions:
% - summarize on high level
% Summary of what we did 
In this project, we focused on developing the theory needed to incorporate equality and functions to the LNN model. We proposed a way to incorporate these elements by implicitly changing the LNN architecture using first-order theories. This allows us to then use equality and functions in the LNNs model proposed by Riegel et al.
The addition of equality and functions allows us to use LNNs to reason about statements in FOL that are much more interesting and natural because we have a larger domain of representable problems. For instance, introduction of equality alone frees us from necessarily working within the unique-names assumption. Additionally, we explain how adding such elements affects the underlying structure of LLNs. With this, we have, therefore, demonstrated that the reasoning ability of LNNs can easily be extended to other domains via first-order theories; moreover, the inclusion of other first-order theories beyond equality and functions would allow LNNs to reason about even more complicated objects.

Future work includes looking into the implementation of additional first-order theories such as theories of arrays, binary search tress, multisets, or even theories of arithmetic. This would then allow one to represent and reason about these structures using a neural network-based architecture. Moreover, future work includes finding more empirical support for the reasoning and learning ability of LLNs by testing their ability to prove theorems on a benchmark set such as TPTP\footnote{\url{https://www.tptp.org/}}; the current implementation of LLNs by Riegel et al. was limited to only a small portion of the theorem proving tasks available because it lacked the ability to represent equality and functions. Additionally, one could also look into LNNs ability to recognize logical entailment as described by \cite{entail} -- something we think LNNs would be naturally well-suited for.

\begin{figure*}[t]
    \centering
    
\begin{fitch}
\fa \fh Pt_1\mydots{}t_i\mydots{}t_n\\
\fa \fa t_i = t_i & $=$-Introduction\\
\fa \fa t_i = t_i \wedge Pt_1\mydots{}t_i\mydots{}t_n & $\wedge$-Introduction via 1, 2\\
\fa \fa \exists x (t_i = x \wedge Pt_1\mydots{}x\mydots{}t_n) & $\exists$-Introduction via 3\\
\fa \\
\fa \fh \exists x (t_i = x \wedge Pt_1\mydots{}x\mydots{}t_n) \\
\fa \fa \exists x (t_i = x \wedge Pt_1\mydots{}t_i\mydots{}t_n) & $=$-Elimination via 6\\
\fa \fa \exists x (Pt_1\mydots{}t_i\mydots{}t_n) & $\wedge$-Elimination via 7\\
\fa \fa Pt_1\mydots{}t_i\mydots{}t_n & $\exists$-Elimination via 8\\
\fa \\
\fa Pt_1\mydots{}t_i\mydots{}t_n \leftrightarrow \exists x (t_i = x \wedge Pt_1\mydots{}x\mydots{}t_n) & $\leftrightarrow$-Introduction via 1-9\\
\end{fitch}\\
    \caption{Rewriting Rule Deduction}
    \label{fig:rulededuction}
\end{figure*}

\newpage

\begin{appendix}

\section{Proof of Function Rewriting Rule}\label{app:funcproof}

Theorem: \ \ \textit{For a given $n$-ary predicate, $P$, and a chosen term $t_i$, $Pt_1\mydots{}t_i\mydots{}t_n$ is logically equivalent to $\exists x (t_i = x \wedge Pt_1\mydots{}x\mydots{}t_n)$.}\\
\vspace{0mm}\\
Proof:

Let $P$ and $t_i$ be arbitrary. As depicted in Figure \ref{fig:rulededuction}, we may deduced that $Pt_1\mydots{}t_i\mydots{}t_n \leftrightarrow \exists x (t_i = x \wedge Pt_1\mydots{}x\mydots{}t_n)$. 
From the Soundess Theorem, it follows that $Pt_1\mydots{}t_i\mydots{}t_n$ is logically equivalent to $\exists x (t_i = x \wedge Pt_1\mydots{}x\mydots{}t_n)$.\qed\\
\vspace{0mm}\\
Corollary:\ \ If $t_i$ is a $k$-ary function, $f(r_1,\mydots{},r_k)$, then $Pt_1\mydots{}t_i\mydots{}t_n$ is logically equivalent to $\exists x (R_fr_1\mydots{}r_kx \wedge Pt_1\mydots{}x\mydots{}t_n)$ where $R_f$ is the $(k+1)$-ary functional predicate associated with $f$.

\iffalse
\subsection{Recursively Extracting Functions}\label{app:recfunc}

To explain the recursive extraction of functions, we do so by providing a basic example of the phenomena. Say we have a unary predicate $P$, two unary functions $f$ and $g$, and a constant $c$. Let $R_f$ and $R_g$ be the functional relations associated with $f$ and $g$, respectively. 

Now say we wish to remove the use of all functions from the formula $Pf(g(c))$. Since $f(g(c))$ is a term of $P$, we first extract this out so that $P$ does not contain any terms which are functions:\[
    Pf(g(c)) \Rightarrow \exists x (R_fg(c)x \wedge Px)
\]
Note that after rewriting, however, we are still left with a predicate containing a term which is a function. While $P$ may be function free, $R_f$ is not. Therefore, we again apply the rewriting rules to our new formula in order to again extract any functions from the terms of our newly introduced predicate $R_f$:\[
    \exists x (R_fg(c)x \wedge Px) \Rightarrow \exists x (\exists y (R_gcy \wedge R_fyx) \wedge Px)
\]
Thus, we now have an equivalent function-free formula for $Pf(g(c))$.
\fi

\end{appendix}

\bibliography{main.bib}

\end{document}